# Hybrid Approach Combining Ultrasound and Blood Test Analysis with a Voting Classifier for Accurate Liver Fibrosis and Cirrhosis Assessment

Kapil Kashyap[1*], Sean Fargose[1], Chrisil Dabre[1], Fatema Dolaria[1], Nilesh Patil[1], Aniket Kore[1]

[1]Dwarkadas J. Sanghvi College of Engineering, Organization, No.U-15, J.V.P.D. Scheme, Bhaktivedanta Swami Marg, Opp.Cooper Hospital, Vile Parle (West), 400056, Mumbai, India.

*Corresponding author(s). E-mail(s): kapilkashyap3105@gmail.com;
Contributing authors: fargose.sean2808@gmail.com;
chrisildabre@gmail.com; fdolaria@gmail.com; nilesh.p@djsce.ac.in;
aniket.kore@djsce.ac.in;

**Abstract**

Liver cirrhosis is an insidious condition involving the substitution of normal liver tissue with fibrous scar tissue and causing major health complications. The conventional method of diagnosis using liver biopsy is invasive and, therefore, inconvenient for use in regular screening. In this paper,we present a hybrid model that combines machine learning techniques with clinical data and ultrasound scans to improve liver fibrosis and cirrhosis detection accuracy is presented. The model integrates fixed blood test probabilities with deep learning model predictions (DenseNet-201) for ultrasonic images. The combined hybrid model achieved an accuracy of 92.5%. The findings establish the viability of the combined model in enhancing diagnosis accuracy and supporting early intervention in liver disease care.

# 1 Introduction

Liver cirrhosis is a chronic and progressive disease whereby the healthy tissues of the liver are gradually replaced by fibrous scar tissue; hence, liver function is reduced significantly. Such conditions are also believed to have multifactorial causes, mainly including autoimmune conditions, NAFLD, alcoholic liver disease with heavy alcoholism, and hepatitis B and C infections. The complications of progressing cirrhosis vary from liver failure, ascites, portal hypertension, and hepatic encephalopathy to hepatocellular carcinoma, which are all considered a set of severe complications with high mortality rates [1]. Liver cirrhosis not only impacts the individual but also brings an immense burden on healthcare systems around the globe.

The correct management and treatment of the patient depend entirely on early detection and proper diagnosis of liver cirrhosis. Early intervention might thereby limit undesirable consequences of the disease and enhance quality of life in affected patients [2]. Diagnosing liver cirrhosis traditionally requires invasive methods, including biopsy of the liver, which is dangerous with a risk of bleeding and infection, as well as being resource-intensive. These limitations make routine screening infeasible, especially in resource-limited settings. Many patients are diagnosed only at an advanced stage of liver disease; poor health outcomes and limited treatment options result.

Non-invasive alternatives include ultrasound imaging and transient elastography as a potential diagnostic tool. The widespread acceptance of ultrasound imaging arises because it is non-invasive; however, at the same time, cost-effective, and can give real-time information about the structure of the liver and blood flow. However, its diagnostic accuracy can be limited, especially in early stages of the disease, where minor changes in the architecture of the liver cannot easily be identified. Transient elastography, which is a method to measure liver stiffness as an indicator of fibrosis, has some promise but has very significant limitations in terms of sensitivity and specificity, especially to differentiate the earliest stage of liver damage from cirrhosis [3].

Despite such promising advancements of ML, however, the existing landscape of diagnosis of liver cirrhosis has several challenges. Most of the existing ML models are designed to target either imaging data or clinical datasets independently, which prevents them from identifying complex patterns and relationships that are critical for making accurate diagnoses. While some models can predict excellently in controlled environments, it ultimately fails to generalize well and determine accurate predictions in real clinical settings, where variability is ubiquitous. This gap spells out the need to have a unified framework which harmonizes clinical data, laboratory results, and imaging modalities for improving diagnostic accuracy and access.

A hybrid model that explicitly leverages the sophistication of machine learning algorithms with multimodal data sources shall complete the gaps for the proposed research. This will develop a consensus diagnostic framework, which shall benefit the non-invasive diagnosis and management of liver cirrhosis in accuracy, reliability, scalability, and adaptability in different clinical settings. The grounds for this research are oriented along the incessant rise in the number of people afflicted with cirrhosis of the liver and its serious health impacts. The silent progress of liver cirrhosis to late stages prior to diagnosis, where only limited options for treatment are viable and complications can subsequently delay recovery, means that their early detection and diagnosis require check-ups. However, the traditional methods like a biopsy are invasive, costly, and entail risks that make them impossible for routine screenings.

Safer alternatives, including blood tests, ultrasound, and other imaging methods have been developed; however, they often fail to distinguish between early liver damage and cirrhosis, at least in diverse clinical environments. Variability in skill of the operator, body composition of the patients, and machine differences often render variable outcomes so that uniform results across several clinics are challenging to achieve.

It appears that the integration of machine learning and artificial intelligence into the diagnostic process may be a promising avenue in overcoming such challenges. Recent studies point out the potential application of ML techniques toward the aim of improving accuracy in detection of liver cirrhosis. For example, algorithms such as SVM and Random Forest have shown powerful performances in classifying the stages of liver disease. The latter produces very high precision and sensitivity and thus is valuable for clinical works [4]. Logistic Regression also performs with good efficacy. It is claimed to achieve over 85% accuracy for different applications in many areas [5].

Advanced models, such as CNNs, also have been employed to further evolve the detection of liver cirrhosis. Given that CNNs are quite successful in image data processing, extraction of subtle features that may hint at liver pathology becomes feasible. Integration of imaging data with clinical and laboratory results provides a deeper understanding of liver health, increasing the accuracy of diagnosis.

The proposed hybrid model aims at improving the deficiency of traditional diagnosis techniques by taking advantage of data sources and making use of highly complex machine learning algorithms. Such a method would improve the precision in the diagnosis of liver cirrhosis besides increasing scalability and adaptability and being suitable for clinical settings in different ways. By minimizing the requirements for invasive procedures, this model will align with the principles of personalized medicine by providing earlier interventions and reducing the burden of liver cirrhosis on healthcare systems.

## 2 Literature Review

The integration of ML and DL techniques has provided improved detection features of liver fibrosis and cirrhosis to diagnose with precise accuracy by ultrasonic imaging with the analysis of blood tests. Artificial Neural Network (ANN)-based approaches have recently been in high demand, where evidence has emerged that these models may be better than the traditional diagnostic techniques, exploiting complex patterns that are not identified by conventional methods, hence increasing the diagnostic accuracy [6]. The inclusion of methodologies and techniques such as fuzzy c-means in the clustering techniques have improved feature extraction from imaging data, thus resulting in improved performance metrics across several studies [7]. Transient elastography has also gained significance, allowing clinicians to integrate imaging results with clinical judgment for a holistic evaluation of liver health [8].

High quality results for ML models have been achieved using diverse datasets, including clinical characteristics, laboratory test results, and imaging data; an example study based on the Kaggle standard dataset reported accuracy rates that range from 76% to over 99% with differing ML algorithms, which strongly suggests that such modeling is quite effective for detecting liver disease [9]. Other feature extraction methods, such as PCA, have been added to modeling, which has increased the predictability models so that they can classify a stage of liver disease [10]. The metrics of accuracy, precision, recall, and F1 score showed good predictability, which indicates their clinical applicability [11].

Interdisciplinary approaches have also been made to know how to handle liver disease in an efficient manner. Wei et al. (2024) introduced LivMarX, a model that simulates liver cirrhosis using biomarkers instead of imaging, employing advanced ML algorithms. This model achieved an accuracy of 84.33% prior to optimization and 86% afterward [12]. This is a big development in early detection and management of liver cirrhosis as ML techniques merge with clinical data and imaging modalities, yielding promising results to the promise that ML can make in the diagnosis of liver disease for better care of patients [13]. Liver cirrhosis is an end-stage manifestation of chronic liver disease and creates serious global health concerns because of complications such as portal hypertension, variceal bleeding, and hepatocellular carcinoma.

While liver biopsy still represents the standard of reference, imaging is widely used in cirrhosis diagnosis and its complications. Traditional morphological criteria employed by imaging have a subjective character and, hence, interobserver variability and diagnostic performance are quite limited. It is with these quantitative imaging methods, which involve artificial intelligence (AI), that better objectivity and diagnostic performance are achieved. Deep learning, especially convolutional neural networks, has revolutionized image analysis across various domains. However, CNNs require huge annotated datasets to train them from scratch, which are scarce in medical contexts.

Transfer learning is effective in adapting a pre-trained CNN to the new task at hand. Nowak et al. (2021) explored how well deep transfer learning can diagnose liver cirrhosis using standard T2-weighted MRI, comparing its performance to that of radiologists with varying levels of experience. The study analyzed data from 713 patients retrospectively, achieving high Dice coefficients for liver segmentation and demonstrating greater accuracy in cirrhosis detection than the radiologists [14]. Furthermore, Endah et al. (2023) improved detection of cirrhosis by incorporating deep learning by using CNN models. It may be used as a potential for the improvement in diagnostic accuracy in cirrhosis [15]. Zhao et al. (2023) experimented with alto-frequency ultrasound images that combined deep learning techniques. Thus, CNN could extract high dimensional features with the enhancement in diagnostic accuracy in cirrhosis [16]. Jabbar et al. (2023) proposed a hybrid ML-based approach that makes use of the ultrasound images of liver fibrosis, achieving a classification accuracy of 98.59% on CNN and SVM classifiers in an ensemble form; indeed, hybrid models improve diagnostic performances[17].

Aggarwal et al. made efficient usage of the Gray Level Co-Occurrence Matrix to extract textural features in ultrasound images. They reported that combining different classifiers and features obtained through the use of the GLCM helped attain a certain level of 89.28% accuracy [18]. In addition, PCA integration for feature extraction has been demonstrated to improve the predictability of models in the classification of stages of liver disease [19]. Recently, efforts have been made to develop automated systems that support clinicians in the diagnosis of liver cirrhosis. For example, research suggested the diagnosis of liver cirrhosis based on altofrequency ultrasound images by using a deep learning-based neural network; the technique enhances the accuracy of diagnosis by extracting high-dimensional features and developing an auxiliary diagnosis system for clinicians [20].

Combining the power of machine learning with clinical data and imaging modalities is a huge step forward in the early detection and management of liver cirrhosis. Promising results from several studies suggest that machine learning is revolutionizing liver disease diagnostics, which should lead to better patient care and outcomes. Future research should target refinement of the already established models, continue hybrid approaches, and try to resolve clinical challenges to make these diagnostic tools better.

## 3 Data Collection and Preprocessing

### 3.1 Clinical Dataset and Preprocessing

This study uses a clinical data set obtained from a Mayo Clinic investigation of primary biliary cirrhosis (PBC) [21]. The data set consists of 418 individual cases of patients with liver cirrhosis. Each case is characterized in terms of 17 clinical features, including demographic information, laboratory test results, and clinical observations. Such features include dimensions such as age, sex, ascites present, enlarged liver (hepatomegaly), and other biochemical parameters such as bilirubin, albumin levels, and prothrombin time. The "Status" variable indicates the status of the survival of the patient as deceased (D), censored (C), or censored because of liver transplantation (CL).

The clinical dataset was subjected to a number of preprocessing operations to make it ready for analysis. Rows with missing values in the "Drug" column were deleted. For other features with missing values, imputation was done using the mean of the corresponding feature. One-hot encoding was done for all categorical attributes to transform them into a numerical format that can be used by machine learning algorithms.

### 3.2 Ultrasound Imaging Dataset and Preprocessing

The ultrasound image dataset is acquired from the research paper in IEEE Access, where the focus was on the liver fibrosis classification from heterogeneous ultrasound images [22]. This dataset consists of images obtained from two South Korean tertiary university hospitals. The total dataset has 6323 images classified into three phases of liver fibrosis, namely No Fibrosis,Fibrosis, and Cirrhosis. They are in JPG format and are used for training and validation of machine learning models intended to classify the level of liver fibrosis.

The ultrasound images were preprocessed to improve their quality and make them ready for analysis. The images were normalized so that brightness and contrast levels remained uniform across the dataset. The images were all resized to the same dimension in order to process them in batches while training the model. Rotation, flipping, and zooming were applied to augment diversity of the training set and model robustness.

## 4 Methodology

### 4.1 Clinical Data Analysis

We used different machine learning algorithms in this study to identify the most optimal model for classifying clinical data related to liver disease, with a focus on fibrosis and cirrhosis presence. The performance of each model was assessed based on accuracy, precision, recall, and F1-score, with separate evaluations conducted for each of the three classes: No Fibrosis (0), Fibrosis (1), and Cirrhosis (2).

LightGBM achieved an accuracy of 0.9533 and was chosen primarily due to its efficient computational efficiency. Its histogram-based methodology reduces memory consumption and increases training time, making it ideal for handling large datasets. The model also showed good precision and recall measures, confirming its overall robust performance.

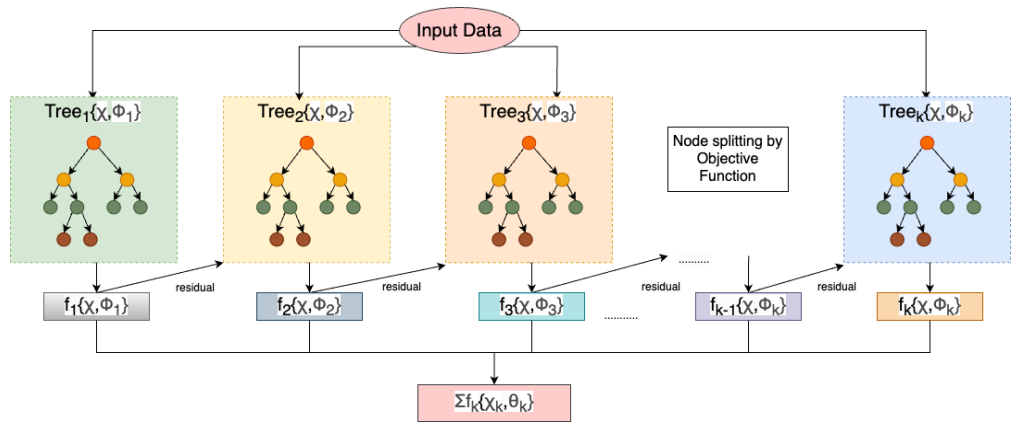

**Fig. 1** XGBoost Architecture Diagram

Above figure depicts the step-by-step workflow of the XGBoost algorithm, including data input, boosting process, and final model output. It illustrates how the algorithm enhances prediction accuracy through iterative training and optimization. XGBoost yielded the top accuracy score of 0.9580, indicating exceptional performance across all three classes. Specifically, it achieved precision scores of 0.96 for No Fibrosis, 0.94 for Fibrosis, and 0.97 for Cirrhosis. Its recall scores were 0.95, 0.96, and 0.96 for the same respective classes, resulting in a consistent F1-score of 0.96 for each. This model's selection was driven by its capacity to efficiently process large datasets, its optimized loss function, and its regularization features designed to mitigate overfitting.

CatBoost reached an accuracy level of 0.9527 and was chosen for its ability to directly manage categorical variables, thereby minimizing the need for extensive preliminary data preparation. The model's precision scores were 0.96 for No Fibrosis, 0.94 for Fibrosis, and 0.97 for Cirrhosis. Similarly, it achieved recall scores of 0.94, 0.96, and 0.96 for the same classes, resulting in uniform F1-scores of 0.95 across all categories.

Random Forest achieved an accuracy of 0.9497 and was employed to mitigate overfitting while providing feature importance insights. This ensemble method constructs multiple decision trees and aggregates their predictions, demonstrating reliable performance through balanced precision and recall metrics.

Gradient Boosting served as a baseline model with 0.8471 accuracy. This stage-wise error-correction approach effectively captured complex data relationships, though it showed higher susceptibility to overfitting compared to modern boosting implementations.

The Gated Recurrent Unit (GRU) achieved 0.9185 accuracy and was investigated for sequential pattern recognition in clinical data. While primarily designed for time-series analysis, its application to potentially sequential clinical data yielded moderate performance below tree-based models.

TabNet attained 0.8981 accuracy using its attention mechanism for interpretable tabular data analysis. This hybrid architecture combines decision tree concepts with deep learning, focusing on clinically relevant features while maintaining model transparency.

Overall, XGBoost emerged as the optimal classifier (accuracy: 0.9580), emphasizing the criticality of model selection for complex clinical data tasks. The superior performance of gradient-boosted tree architectures highlights their effectiveness in liver disease classification.

## 4.2 Ultrasound Imaging Analysis

We used a number of deep learning methods to classify ultrasonic image data with the aim of identifying important features for precise diagnosis. The models tested are AlexNet, DenseNet, ResNet50V2, and a typical Convolutional Neural Network (CNN).

AlexNet had an accuracy of 0.7928. This groundbreaking convolutional neural network has five convolutional layers with max-pooling layers and uses ReLU activation functions to provide non-linearity. AlexNet is capable of dealing with large sets of images and employs dropout and data augmentation techniques to prevent overfitting. Although its performance on ultrasonic images is satisfactory, it may be limited compared to recent architectures.

ResNet50V2 had an accuracy of 0.7895. The model uses residual connections that make it easier to train deeper networks and overcome the vanishing gradient issue. ResNet50V2 facilitates the construction of networks of hundreds of layers with high accuracy. In ultrasonic imaging, ResNet50V2 can identify complex features and patterns but at lower accuracy than DenseNet.

CNN (Convolutional Neural Network) performed with an accuracy of 0.8213. A standard CNN consists of many convolutional layers and pooling layers which are designed to learn spatial hierarchies of features automatically from images. CNNs are very versatile and can be configured for particular tasks, making them very popular for image classification. For ultrasound images, CNNs can be trained to prioritize significant features, achieving similar accuracy.

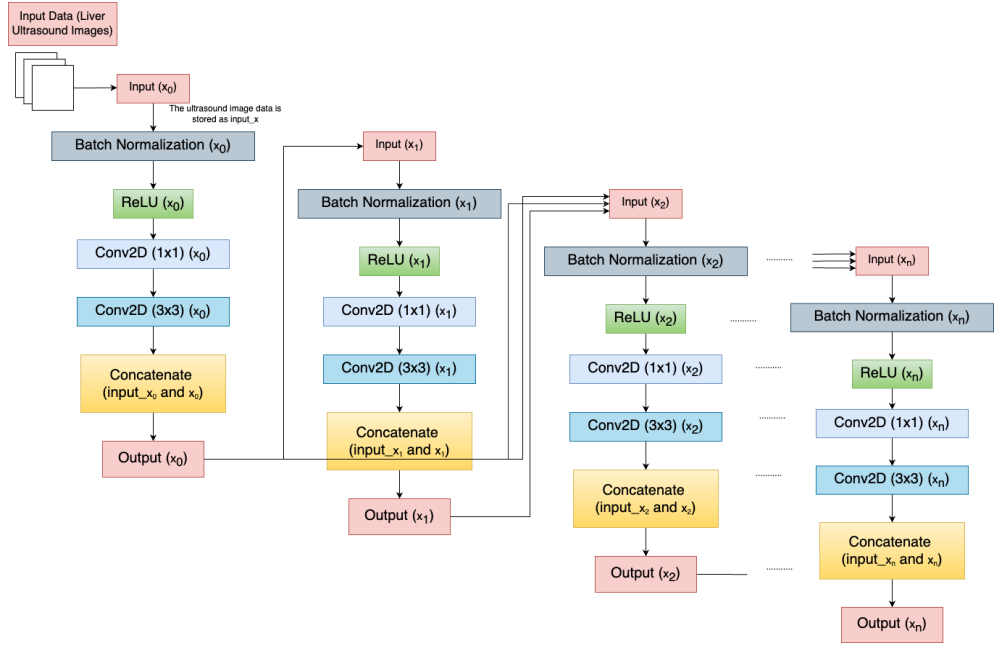

**Fig. 2** DenseNet Architecture Diagram

Above figure shows a multi-path convolutional neural network structure that is optimized to process liver ultrasound images. The network starts with the input of the ultrasound image data, denoted as (input_x). All branches of the network have batch normalization applied to them followed by a ReLU activation function.The paths consist of two sequential convolutional layers in order to allow the network to extract various features from the images.

After each sequence of convolutional layers, the results are concatenated with the initial input (input_x), providing a richer feature representation.This concatenation happens twice throughout the architecture, showing the capability of the network to concatenate low-level features with higher-level abstraction well.Once completed, each branch is finished by an output, allowing extra processing or classification procedure. In total, this structure focuses on depth and feature combination, important for precise image analysis in medical practice.

DenseNet achieved a maximum accuracy of 0.91 among the models considered. DenseNet, or Densely Connected Convolutional Networks, favors feature propagation and reuse by making each layer connected with all the other layers in a feed-forward fashion. The design minimizes the number of parameters but maximizes accuracy and efficiency. DenseNet's capacity to learn from numerous features is especially efficient in complex image classification problems, such as ultrasonic image data.

Overall, DenseNet was the top-performing model for ultrasonic image data classification in this study, illustrating the importance of selecting appropriate deep learning architectures for complex image classification tasks in medical imaging. The differences in model performance highlight the critical role of careful architectural selection in maximizing diagnostic accuracy.

## 5 Hybrid Predictions using Voting Classifiers

This work proposes a hybrid diagnostic classifier system that uses predictions from the analysis of blood tests and ultrasound images and fuses them using voting classifiers. It enhances the reliability and accuracy of the detection of liver fibrosis and cirrhosis by leveraging the respective strengths of the two diagnostic modalities. It dynamically weighs each modality based on their respective classification accuracies and produces a well-calibrated decision process. The method uses a soft voting classifier, which averages out probabilistic output of multiple classifiers and returns the final diagnosis by selecting the class with highest cumulative probability, thereby allowing more precise predictions to make stronger influence on the outcome.

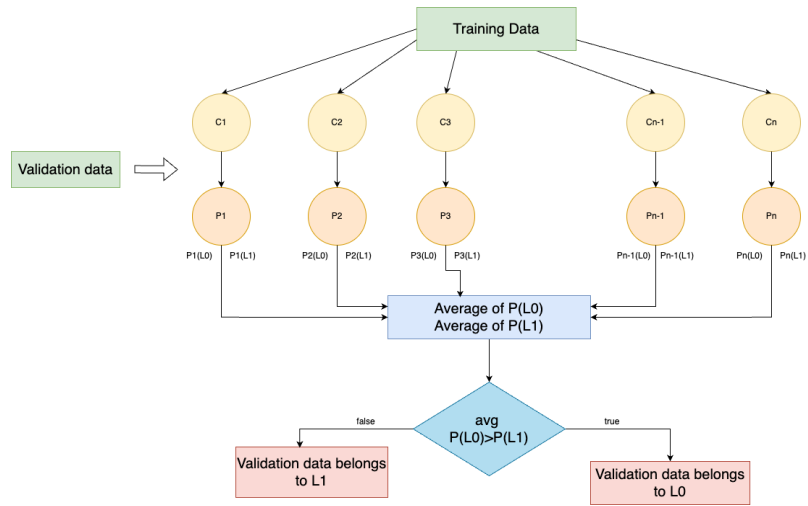

**Fig. 3** Architecture of a Soft Voting Classifier

The above figure shows the architecture of a soft voting classifier for binary classification in two classes L1, L2 using n individual classifiers ($C_1, C_2, \ldots, C_n$), where there are several classifiers $C_1, C_2, \ldots, C_n$ trained from the data and each classifier generates individual probability distributions $P_1, P_2, \ldots, P_n$ for every class. The final classification is attained by computing the average probabilities across all the classifiers, giving the validation data to the most cumulative likely class enhancing the robustness and reliability of the decision-making process.

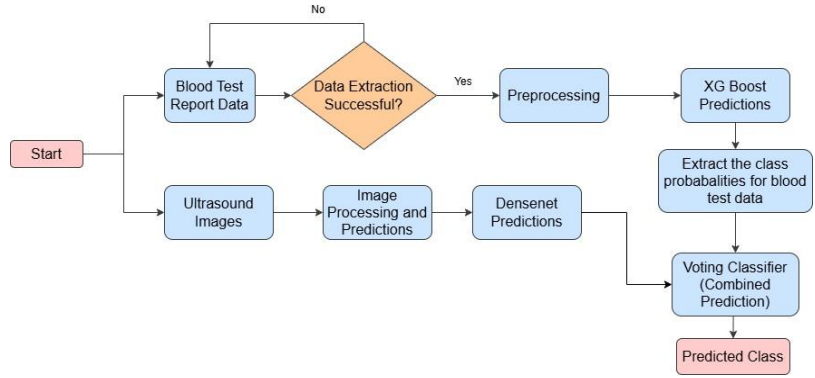

**Fig. 4** Workflow diagram for Soft Voting Classifier

The process of the hybrid diagnostic model for detection of liver fibrosis and cirrhosis to be proposed is illustrated in the diagram. The process starts with drawing blood test data and ultrasound images. On success, the blood test data is preprocessed and tested through an XGBoost model, whereas ultrasound images are processed and classified in a DenseNet model. The probability scores from both modalities are then weighted according to their classification accuracy and aggregated with a soft voting classifier to arrive at the final predicted class (No Fibrosis, Fibrosis, or Cirrhosis). If data extraction is unsuccessful, the process stops. The weighting factor for the blood test data, $w_b$, is determined using the equation:

$$w_b = \frac{A_b}{A_b + A_i} \quad (1)$$

Similarly, the weighting factor for image-based analysis $w_i$ is given by:

$$w_i = \frac{A_i}{A_b + A_i} \quad (2)$$

where,

- $w_b$ = Weighing factor for the blood test model
- $w_i$ = Weighing factor for the blood test model
- $A_b$ = Accuracy of the proposed model for blood test data
- $A_i$ = Accuracy of the proposed model for image analysis model

These weights are then multiplied against the corresponding class probabilities of the individual models. The final class probability in the hybrid model is calculated as the weighted sum of the probabilities from the individual models. This provides more important contributions in the classification decision in the final step by the models having a higher level of accuracy.

**Equation for Hybrid Probability Calculation**

The final hybrid probability for class *c* can be computed as:

$$P_{\text{hybrid},c} = w_b \cdot P_{b,c} + w_i \cdot P_{i,c} \qquad (3)$$

where,

- $P_{b,c}$ = Predicted probability of class *c* from the blood test model
- $P_{i,c}$ = Predicted probability of class *c* from the image analysis model
- $P_{\text{hybrid},c}$ = predicted probability of hybrid model for class *c*

This method sets up an equitable and strong prediction system, which guarantees that both structured (clinical) and unstructured (image) data significantly contribute to the final diagnosis. The weighted voting process overcomes individual model weaknesses to allow for better and more complete evaluation of liver fibrosis and cirrhosis.

# 6 Results

Several machine learning models were evaluated in this study for their ability to classify clinical data pertaining to liver conditions. Model performance was gauged using accuracy, precision, recall, and F1-score, with individual evaluations conducted for each of the following three classes: No Fibrosis (0), Fibrosis (1), and Cirrhosis (2). The results are summarized in Tables 1 and 2.

**Table 1** Accuracy and Precision for Clinical Data Analysis

| Model | Accuracy | Precision (0) | Precision (1) | Precision (2) |
|---|---|---|---|---|
| XGBoost | 0.9580 | 0.96 | 0.94 | 0.97 |
| LightGBM | 0.9533 | 0.95 | 0.94 | 0.97 |
| CatBoost | 0.9527 | 0.96 | 0.94 | 0.97 |
| Random Forest | 0.9497 | 0.95 | 0.93 | 0.97 |
| Gradient Boosting | 0.8471 | 0.85 | 0.94 | 0.97 |
| GRU | 0.9185 | 0.93 | 0.89 | 0.94 |
| TabNet | 0.8981 | 0.90 | 0.87 | 0.91 |

**Table 2** Evaluation matrix for clinical data

| Model | Recall (0) | Recall (1) | Recall (2) | F1 (0) | F1 (1) | F1 (2) |
|---|---|---|---|---|---|---|
| XGBoost | 0.95 | 0.96 | 0.96 | 0.96 | 0.95 | 0.97 |
| LightGBM | 0.95 | 0.95 | 0.96 | 0.95 | 0.95 | 0.97 |
| CatBoost | 0.94 | 0.96 | 0.96 | 0.95 | 0.95 | 0.96 |
| Random Forest | 0.94 | 0.95 | 0.96 | 0.94 | 0.94 | 0.97 |
| Gradient Boosting | 0.83 | 0.96 | 0.96 | 0.84 | 0.95 | 0.97 |
| GRU | 0.91 | 0.92 | 0.93 | 0.92 | 0.90 | 0.93 |
| TabNet | 0.90 | 0.89 | 0.91 | 0.90 | 0.88 | 0.91 |

The results show that XGBoost achieved the best accuracy (0.9580), supported by high precision, recall, and F1-scores for all classes. LightGBM and CatBoost also performed well, with Gradient Boosting performing relatively lower accuracy and F1-scores. The results reflect the effectiveness of the models being tested in classifying clinical data for liver disease, and how model choice contributes to achieving best performance results.

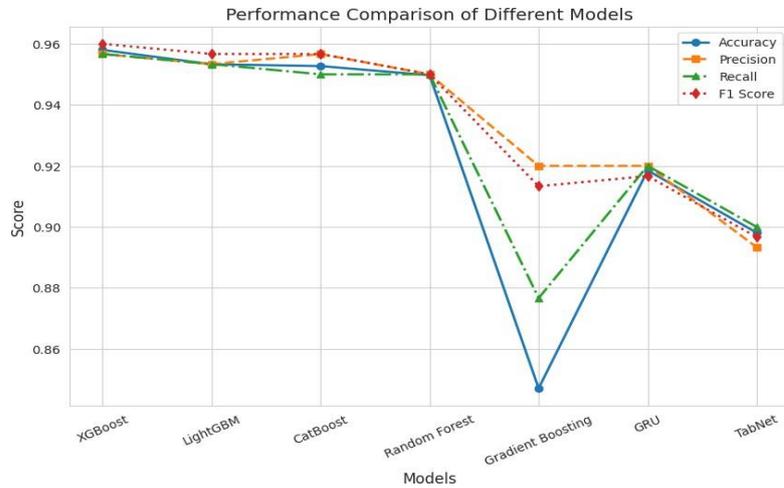

**Fig. 5** Graphical Representation of Clinical Data Results

**Table 3** Accuracy for Ultrasonic Image Dataset

| Model | Accuracy |
|---|---|
| AlexNet | 0.7928 |
| DenseNet | 0.9071 |
| ResNet50V2 | 0.7895 |
| CNN | 0.8213 |

The accuracy results of the deep learning models, as evaluated on the ultrasonic image dataset, are presented and summarized in Table 3.

These results demonstrate the effectiveness of the deep learning models assessed in classifying ultrasonic image data.

The hybrid model, integrating clinical and image data, was evaluated on a cohort of 40 patients. This model incorporated blood test probabilities in conjunction with predictions generated by a DenseNet-201 model analyzing ultrasonic images.

This hybrid model yielded a successful accuracy of 92.5%, demonstrating the effectiveness of integrating clinical and image data to enhance diagnostic accuracy.

# 7 Future Scope

Future work can then continue to evolve the hybrid model to take in larger multimodal input data, for example, genomic data, lifestyle data, and demographic data, to enable more accurate diagnostics and personalization. The application of sophisticated feature selection methods will also enable the determination of the most discriminative predictors of liver disease. Furthermore, increasing the dataset size even larger and from a more heterogeneous population can offer the model more generalizability and robustness across various clinical practices. In addition, real-time application of the model in practice may give timely decision-making and enhance patient outcomes in the management of liver disease.

# 8 Conclusion

Blending the clinical data and the ultrasound pictures through a hybrid machine learning technique is an impressive advance in the non-invasive diagnosis of cirrhosis and liver fibrosis. By merging the advantage of blood test analysis and deep learning techniques, the model acquired a high accuracy rate of 92.5%, which demonstrates its capability in real-world applications. It not only improves the accuracy of diagnosis but also follows the personalized medicine approach by facilitating the treatments at an earlier stage. Subsequent research would be tuning the model to its optimal level and testing its performance across different clinical settings in order to render it generalizable and effective in practice. The findings indicate the importance of creating hybrid models in the war against liver disease, resulting in improved patient outcomes and reduced healthcare expenses.